\documentclass{article}

\usepackage{PRIMEarxiv}

\usepackage[utf8]{inputenc} 
\usepackage[T1]{fontenc}    
\usepackage{hyperref}       
\usepackage{url}            
\usepackage{booktabs}       
\usepackage{amsfonts}       
\usepackage{nicefrac}       
\usepackage{amsmath}
\usepackage{microtype}      
\usepackage{lipsum}
\usepackage{fancyhdr}       
\usepackage{graphicx}       
\usepackage{caption}
\graphicspath{{media/}}     

\title{CLIP-Diffusion-LM: Apply Diffusion Model on Image Captioning
}

\author{
  Shitong Xu \\
  Imperial College London \\
  \texttt{shitong.xu19@imperial.ac.uk} \\
}

\begin{document}
\maketitle

\begin{abstract}
Image captioning task has been extensively researched by previous work. However, limited experiments focus on generating captions based on non-autoregressive text decoder. Inspired by the recent success of the denoising diffusion model on image synthesis tasks, we apply denoising diffusion probabilistic models to text generation in image captioning tasks. We show that our CLIP-Diffusion-LM is capable of generating image captions using significantly fewer inference steps than autoregressive models. On the Flickr8k dataset, the model achieves 0.1876 BLEU-4 score. By training on the combined Flickr8k and Flickr30k dataset, our model achieves 0.2470 BLEU-4 score. Our code is available at \href{https://github.com/xu-shitong/diffusion-image-captioning}{https://github.com/xu-shitong/diffusion-image-captioning}.

\end{abstract}


\section{Introduction}
Image captioning has been a focus of research over recent years. Among the previous proposed works, the text encoder used can be classified into 2 general classes, i.e. autoregressive and non-autoregressive class. Most of the state-of-the-art models fall in the autoregressive class \cite{https://doi.org/10.48550/arxiv.1412.6632,DBLP:journals/corr/XuBKCCSZB15,DBLP:journals/corr/LuXPS16,image-caption-with-pos,mplug}. However, autoregressive generation suffer from 1) the slow generation speed due to the generation step is token by token, and 2) not capable of refining prefix of sentences based on the later generated tokens. Multiple attempts have experimented using a non-autoregressive model in the text generation steps \cite{masked-non-autoregres, partial-autoregressive, semi-autoregressive}. The closest work to ours is Masked Non-Autoregressive Image Captioning by Gao et al.  \cite{masked-non-autoregres}, which uses a BERT model as the generator and involves 2 steps-refinement on the generated sequence. Masked Language Model (MLM) is used in their work to supervise the generation of captions.

In contrast to MLM, which is a language model based on discrete tokens embedding prediction, diffusion models based on continuous latent embedding have been thriving in image and audio generation tasks \cite{dalle2, https://doi.org/10.48550/arxiv.2205.11487,glide,DBLP:journals/corr/abs-2105-06337,https://doi.org/10.48550/arxiv.2207.09983}. To the best of our knowledge, there has not been previous work on generating caption embedding based on a diffusion language model. Our work aim at employing a model to refine generated token continuously on sequence embedding, and provide empirical insight on useful tricks to improve the generated captions. In particular, we use pre-trained CLIP \cite{clip} model for extracting image and text features, and DistilBert \cite{distilbert} model based on Diffusion-LM \cite{diffuselm} for text sequence generation. Our contribution consists of proposing a diffusion-based image captioning model (\hyperref[sec:CLIP-DiffuseLM]{Sec.4}), and experiments on effectiveness of multiple model design and hyperparameter settings, including classification free guidance in \hyperref[sec:classification-free-exp]{Sec.5.1}, learning rate in \hyperref[sec:lr-exp]{Sec.5.2}, weight assign in loss function terms in \hyperref[sec:lambda-exp]{Sec.5.3}, $x_0$-prediction in \hyperref[sec:x0-exp]{Sec.5.4} and feature fusion method in \hyperref[sec:CLIP-DiffuseLM]{Sec.5.5}.

\section{Related Work}
\label{sec:headings}

\subsection{Autoregressive image captioning}
Mao et al. proposed the mRNN model \cite{https://doi.org/10.48550/arxiv.1412.6632}, which uses CNN for image extraction and RNN for text generation. Xu et al.  \cite{DBLP:journals/corr/XuBKCCSZB15} applied the LSTM to text generation and experimented on soft and hard attention for early fusion between image feature and text feature. Based on this early fusion method, Lu et al. \cite{DBLP:journals/corr/LuXPS16} experimented with the late fusion of image and text features, allowing model to attend on either image or text modality in the late phase of generation. Wu and Hu \cite{cascade-RNN} experimented on reversing the generated caption, allowing its backend model to refine the former tokens based on later caption tokens. Le et al. \cite{gla} used attention model to combine local and global features from images, so that captions can more accurately identify occluded objects. Similarly, Wei et al. \cite{stack-vs} also used image features from both high and low generality, and combined them using cross-attention. Their work also involves multi-step refining of the generated text caption. Feng et al. \cite{DBLP:journals/corr/abs-1811-10787} trained image caption in a GAN framework, with a LSTM discriminator reproducing the original image feature from generated text sequence. Similarly, Guo et al. \cite{mscap} proposed GAN based method to train model predicting stylized text. Multiple discriminators are used to supervise if generated text captured image-related feature, is in the desired style, and is similar to a caption made by human. Kim et al.  \cite{vae-caption} used variational autoencoder for extracting image information, their model allows multi-caption generation by sampling from the learned image feature distribution, thus produce various captions for a single image. He et al. \cite{image-caption-with-pos} used POS tagging to help the generation of text. The image feature is used as additional input when the model is predicting tokens related to image-specific information, i.e. object, colour, relative position of objects. Mokady, Hertz and Bermano \cite{clipcap} experimented with using a pre-trained CLIP image feature for sequence generation. The CLIP features are transformed to a sequence of token and used as prefix for a GPT-2 model in generation. Li et al. \cite{mplug} introduced skipped connections between transformer layers to address the information asymmetry between vision and language modality. The model achieves state-of-the-art performance and strong zero-shot ability on various tasks. Nguyen et al. \cite{grit} experimented with changing the cross attention part of transformer decoder to use both Regional feature from Faster-RCNN \cite{faster-rcnn} and Grid features from Swin transformer \cite{swin}. 

\subsection{Non autoregressive image captioning}
In contrast, non-autoregressive models benefit from the attention models' ability to pass textural information in both directions during generation. The text generated in former timesteps could adjust based on text in later timesteps, thus is expected to achieve better performance. Gao et al. \cite{masked-non-autoregres} used BERT \cite{bert} as text decoder and employed a 2 step generation method. Based on this work, Partially Non-Autoregressive Image Captioning by Fei \cite{partial-autoregressive} and semi Non-Autoregressive Image Captioning by Xu et al. \cite{semi-autoregressive} partitioned the generated text in subgroups. Words in the same group are generated non-autoregressively and different groups are generated autoregressively. Our model falls in non-autoregressive category and is most close to the Masked Non-Autoregressive Image Captioning \cite{masked-non-autoregres}. The difference is we choose to use the diffusion model as the non-autoregressive generation model.

\subsection{Diffusion models}
Diffusion model aims at training a model that denoise Gaussial noise incrementally to reproduce original features. Ho, Jain and Abbeel \cite{ddpm} proposed the Denoising Diffusion Probabilistic Model (DDPM) to simplify the loss function by only letting models predict the noise in generation steps, and proposed an alternative loss function by removing the weight coefficients. In the following explanation, we refer to diffusion model as DDPM for simplicity. Nichol and Dhariwal \cite{improved-ddpm} proposed several improvements based on DDPM, including setting variance to be learn-able parameters, apply cosine instead of linear noise schedule, and speed up forward process by reducing forward steps. Song, Meng and Ermon \cite{ddim} experimented on reducing the variance in forward process. The result shows that by reducing variance to 0, the deterministic model achieves higher FID score in image generation on both CIFAR10 and CelebA. Diffusion-LM by Li et al. \cite{diffuselm} is a recent work on applying continuous diffusion model on text generation. Their work explored various techniques to improve the performance of continuous diffusion model on text generation. 

Dhariwal and Nichol \cite{diffuseBeatGan} proposed classifier guidance for improving generated image FID score. In a classifier-guided diffusion model, a classifier model is pretrained to predict noised images' object class. During training, the classifier provides gradient on which direction to optimise the generated image, so that the generate image resembles an object closer to the target class. 

To avoid training classifier for guiding model, Jonathan and Tim  \cite{classifier-free} proposed classifier-free guidance model. In classifier-free guidance, the difference between outputs of generative model when provided with either guided and unguided context information is used as implicit guidance. By using classifier-free diffusion model as text-to-image generator, DALL-E2 \cite{dalle2}, GLIDE \cite{glide} and High-Resolution Image Synthesis With Latent Diffusion Models \cite{high-resolution-image-synthesis} model achieves significant image generation performance. In particular, DALL-E2 use CLIP model for extracting feature from text, predict the corresponding image CLIP feature through prior network, then use predicted image CLIP feature for final image generation. The model achieves significant novelty in generated images and also inspired us to train a image-to-text model with diffusion model in generation step.

\subsection{CLIP model}
CLIP model is a contrastive-learning-based model trained on WebImageText dataset. The WebImageText consists of 400 million image-text pairs collected from publicly available sources on the Internet. CLIP model demonstrates strong zero-shot performance in their evaluation on ImageNet \cite{imagenet} dataset. Further work on CLIP model also shows its transferability to other tasks, including image segmentation \cite{lseg, groupvit}, objection detection \cite{vild, glip}, action recognition \cite{clip4clip} and video-text retrieval \cite{actionclip}. 

\section{Background}

\subsection{Diffusion models}
The training of the denoise diffusion probabilistic model involves generation of noised samples (forward process), and the model denoising Gaussian noise back to the original feature (backward process). Let $x_t$ represent the intermediate representation between the diffusion generation stages. In our context, $x_t$ is the sequence of word embedding of the caption. The forward process incrementally adds noise to the ground truth embedding $x_0$ to generate T noised features $[x_1, ... x_T]$. Each $x_t$ at step t is sampled from probability distribution $q(x_t | x_{t-1}) = N(x_t; \sqrt{1 - \beta_t}x_{t-1}, \beta_t I)$ and the final step feature $x_T$ is aiming at approximate a Gaussian noise. From reparameterization trick, the $x_t$ at any step t could be directly sampled from $x_0$: $x_t = \sqrt{\bar{\alpha}_t}x_0 + (1 - \bar{\alpha}_t) \epsilon$, where $\bar{\alpha}_t = \prod_{s = 1}^t(1 - \beta_s)$ and $\epsilon$ follows a Multivariate Gaussian Normal distribution. The backward process involves training model with parameter $\theta$ to denoise the samples generated in the forward process. The training objective is to minimize the negative log-likelihood of generating $x_0$ from arbitrary Gaussian noise $x_T$ as generated by $q$, that is to minimize
\begin{align}
    E_q[-\log(p_{\theta}(x_0))] &= E_q[-\log(\int p_{\theta}(x_{0:T}) d(x_{1:T}))]\,.
\end{align}
By optimizing the variational lower bound of $p_{\theta}(x_0)$ instead, and modeling the backward process as a Markov Process gives:
\begin{align}
    E_q[-\log(p_{\theta}(x_0))] &\leq E_q[\log(\frac{p_{\theta}(x_{0:T})}{q(x_{1:T}| x_0)})] \\
    &= E_q[\log(p(x_T)) + \sum_{t = 1}^T\log(\frac{p_{\theta}(x_{t-1} | x_t)}{q(x_t | x_{t-1})})]
\end{align}

where $p_{\theta}(x_{t-1} | x_t) = N(x_{t-1}; \mu_{\theta}(x_t, t), \sum_{\theta}(x_t, t))$ and $\mu_{\theta}(x_t, t)$ is model's prediction on mean of $x_{t-1}$ conditioned on $x_t$. From DDPM \cite{ddpm}, expanding and reweighting each term of the negative log-likelihood gives a concise loss function 
\begin{align}
    L_{simple} = \sum_{t=1}^T E_{q(x_t | x_0)} \|\mu_{\theta}(x_t, t) - \hat{\mu}(x_t, x_0)\|^2
\end{align}
where $\hat{\mu}(x_t, x_0)$ is the mean of posterior $q(x_{t-1} | x_t, x_0)$.

Due to the large generation step number (T = 1000 as proposed in  \cite{ddpm}), and the generation step being autoregressive on the denoised feature in the previous step, the reverse diffusion is significantly slower than the other generative models \cite{gan, vae}. Multiple strategies were proposed to accelerate the generation process. Improved DDPM proposed by Nichol and Dhariwal \cite{improved-ddpm} uses a subset of generation steps $\{s_0, ..., s_N | s_t < s_{t-1}\} \in (0, T]$. Model is trained to predict $x_{s_{t-1}}$ based on $x_{s_t}$. Diffusion-LM proposed by Li et al. \cite{diffuselm} is trained directly to predict the $x_0$ instead of an denoised intermediate steps $x_{t-1}$.

In addition, we instead use a L1 loss and add a rounding term and a $x_1$ restoring loss term to the loss function as propose by Li et al. \cite{diffuselm}. $x_1$ restoring loss $\|\mu_{\theta}(x_1, 1) - \hat{\mu}(x_1, x_0)\|$ evaluates the performance of model on restoring $x_1$. For simplicity, in the following explanation and experiments, we evaluate the $x_1$ restoring loss together with $L_{simple}$ and refer to their sum as $L_{simple}'$. Rounding term $L_R$ is parameterized by $E_q[-\log(p_{\theta}(w | \hat{x}))] = E_q[-\log(\prod_{i=1}^l p(w_i | \hat{x}_i))]$, where $l$ represents the generated sequence length, $w$ represent the ground truth sentence and $\hat{x}$ is the predicted sequence embedding from the input $x_t$. $p_{\theta}(w_i | \hat{x}_i)$ follows the Softmax distribution. In our experiments, we find the relative importance between rounding term $L_R$ and restoring embedding loss $L_{simple}'$ significantly influence the model's performance. The relative importance is addressed by the hyperparameter $\lambda$ as rounding term coefficient and discussed in detail (\hyperref[sec:lambda-exp]{Sec.5.3}). Based on the above modifications, our training objective is as follow:

\begin{align}
    L_{diffuseLM} = L_{simple}' + \lambda L_R = \sum_{t=1}^{N} E_{q(x_{s_t} | x_0)} [\|\mu_{\theta}(x_{s_t}, s_t) - \hat{\mu}(x_{s_t}, x_0)\| -\lambda\log(p_{\theta}(w | \hat{x}))]\,.
\end{align}

\section{CLIP-Diffusion-LM}
\label{sec:CLIP-DiffuseLM}
\begin{figure}
    \centering
    \includegraphics[width=\textwidth]{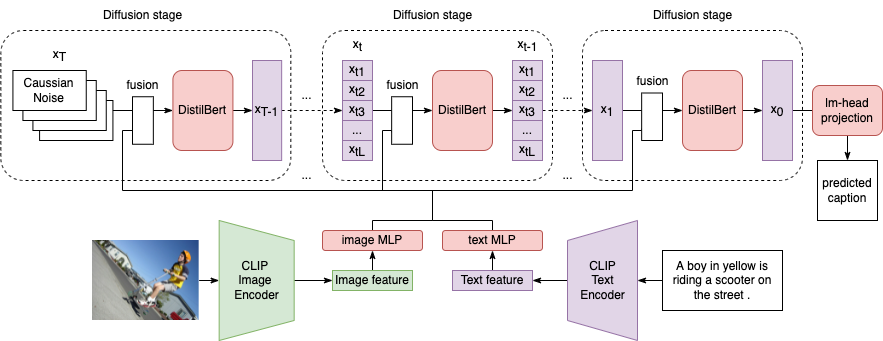}
    \caption{CLIP-Diffusion-LM model}
    \label{fig:model}
\end{figure}

Our CLIP-Diffusion-LM model (Figure 1) consists of 2 parts. A pre-trained CLIP (ViT-B/32) model \cite{clip} is used for image and ground truth caption label feature extraction, and a DistilBert-base-uncased model \cite{distilbert} for performing denoising diffusion in each diffusion stage conditioned on CLIP image feature.

In a forward process, the CLIP model first extract image features using a ViT-Base/32 architecture \cite{vit}. In this model, source RGB image is partitioned into 32 patches and prefixed with a <CLS> token to form the input of the 12 ViT layer model. Output feature at the <CLS> token position is treated as the global feature for the whole image. For classification free experiments (\hyperref[sec:classification-free-exp]{Sec.5.1}), CLIP text feature is also extracted as guidance context to improve the performance. Model used in CLIP caption feature extraction is a modified transformer encoder model \cite{gpt, transformer}. Activation at the highest transformer layer in <EOS> token is used as the global caption feature. In our following experiments, both CLIP text and image branches parameters are not optimized in the backward propagation.

The generation of captions for a given image consists of a sequence of diffusion stages conditioned on CLIP feature extracted. In each stage, CLIP-Diffusion-LM receives input of $L \times D_{word}$ dimension vectors $x_t$ as embedding of a $L$ length caption, and two $D_{CLIP}$ dimension vector for CLIP text and image features. Input caption embedding is either the output embedding $x_t$ in the previous diffusion stages, or the Gaussian noise $x_T$ in the case of the first diffusion stages. Each of CLIP features is projected by a MLP layer to $D_{word}$ space, before fuse (\hyperref[sec:fusion-exp]{Sec.5.5}) with the sequence embedding $x_t$. The fused sequence is the input of the DistilBert model, and output of the DistilBert model is the output of one single diffusion stage, which consists of the prediction for $x_{t-1}$. After the final diffusion stage, the model's prediction for $x_0$ is linear projected by weight matrix \texttt{lm-head} and taken the Softmax value to get the probability of the predicted word in each $L$ length caption sequence position. Finally, each position's word with maximum probability are concatenated to form the output text caption sequence. Unless otherwise specified, both the embedding layer for extracting ground truth caption embedding and the \texttt{lm-head} use pre-trained DistilBert model embedding layer parameters and are not optimized in model training.

\section{Experiments}
We train and evaluate our baseline model on Flickr8k \cite{flickr8k} dataset, which consists of 8k images and 5 captions for each image. After adjusting learning rate (\hyperref[sec:lr-exp]{Sec.5.2}), we train our final model on combined dataset of Flickr30k \cite{flickr30k} and Flickr8k \cite{flickr8k}. The combined dataset, which we refer to as Flickr30+8k dataset, consists of 38k images and 190k captions. Both models early stop when validation loss is higher than training loss, and result in 15 epochs training used for both models.

The baseline model uses configuration of batch size of 8, max output sequence length 16, $\lambda = 0.3$, learning rate of $5e-5$, concatenation fusion and non-classification-free guidance. It takes around 5 hours to train the baseline model for 15 epochs using AdamW optimizer on a single Nvidia A30 GPU. During training, a subset of 100 $x_t$ noise embeddings are randomly sampled from the total 1000 intermediate representations. The model is trained to predict mean $x_0$ directly as suggested in Diffusion-LM \cite{diffuselm}. In the final inference step, the baseline model refines its output autoregressively for 5 iterations, and all but the first element from every consecutive group of repeated words are removed to improve the BLEU-4 performance. The final model uses the same configuration as the baseline model, apart from decaying learning rate linearly from $1e-4$ to $5e-5$. The final model takes around 11 hours to finish 10 epochs training on Flickr30+8k dataset.

\subsection{classification free guidance}
\label{sec:classification-free-exp}
The effectiveness of applying classification-free guidance is examined in our model. The guidance provided is the CLIP text feature of ground truth label caption. Table 1 shows the comparison between the baseline and classification-free guidance-trained model. In contrast to the previous success of applying classification-free guidance, our model fails to improve significantly over none classification guided baseline. We further test the 2 classification-free-guided models' performance by evaluating BLEU score on the validation set. The model trained using classifier-free guidance hardly outperform the simpler baseline model, with the parameter suggested by Jonathan and Tim  \cite{classifier-free} ($w = 0.3$, $p_{uncond} = 0.2$) even result in slight decrease in BLEU-4 score. As a result, classification-free guidance is not used in our final model.

\begin{minipage}[c]{0.49\textwidth}
    \centering
    \begin{tabular}{lllll}
    \toprule
    $w$ & $p_{uncond}$ & $L_{simple}'$ & $L_R$ & BLEU-4 \\
    \midrule
    - & - & 4.89 & 12.87 & 0.1549 \\
    0.3 & 0.2 & 4.92 & 13.49 & 0.1539 \\
    1.0 & 0.2 & 4.89 & 13.06 & 0.1558 \\
    \bottomrule
    \end{tabular}
    \label{tab:classifier-free}
    \captionsetup{type=table}\captionof{table}{classifier-free guidance performance comparison, first row represents non-classifier-free guidance performance}
\end{minipage}
\begin{minipage}[c]{0.5\textwidth}
    \centering
    \begin{tabular}{llll}
    \toprule
    $lr$ & $L_{simple}'$ & $L_R$ & BLEU-4 \\
    \midrule
    5e-5 & 4.89 & 12.87 & 0.1549 \\
    1e-4 & 4.73 & 11.11 & 0.1699  \\
    log schedule & 4.82 & 11.30 & 0.1643 \\
    linear schedule & 4.82 & 11.28 & 0.1876 \\
    cosine annealing & 4.79 & 11.28 & 0.1848 \\
    \bottomrule
    \end{tabular}
    \label{tab:lr}
    \captionsetup{type=table}\captionof{table}{BLEU-4 score of different learning rates}
\end{minipage}

\subsection{learning rate}
\label{sec:lr-exp}
\begin{figure}
  \centering
  \includegraphics[width=\textwidth]{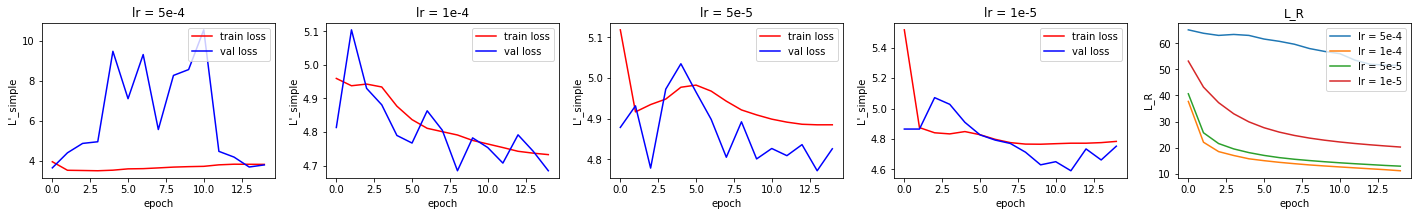}
  \caption{constant learning rate loss}
  \label{fig:lr}
\end{figure}

We experiment on constant learning rate of $lr \in \{5e-4, 1e-4, 5e-5, 1e-5\}$ (Figure 2 and Table 2). Among these learning rate choices, $lr = 1e-4$ achieve the highest BLEU-4 score (0.1699), and the lowest $L_{simple}'$ loss (4.73). However, the model fails to converge to both a lower $L_{simple}'$ and $L_R$ loss before over-fitting when the learning rate is set below $1e-4$. A probable cause is training with a low learning rate prevents the model from exploring the parameter space and converging to a local minimum.

To experiment on if model can refine the result achieved by a larger learning rate, we apply linear, log, and cosine annealing scheduling between the 2 best-performing learning rate values ($1e-4$ and $5e-5$). Among the 3 scheduling methods, linear scheduling achieve the highest BLEU-4 score (0.1876), followed by cosine annealing (0.1848). Log scheduling, which decreases learning rate significantly more rapidly to a low value, fail to achieve comparable performance to the other 2 scheduling method. This failed experiment also supports our guess above. In addition, when extending the dataset to use Flickr30k dataset, both the final model trained on Flickr8k and Flickr30+8k dataset early stop before 15 epochs (12 epochs for Flickr8k and 10 epochs for Flickr30+8k) compared to the other experiments' 15 epochs, indicating linear learning rate decay between $1e-4$ and $5e-5$ helps model converge faster. As a result, linear decay of learning rate from $1e-4$ to $5e-5$ is chosen as our final model configuration. 

\subsection{Relative importance between Rounding term \texorpdfstring{$L_R$}{TEXT} and Embedding-restoring loss term \texorpdfstring{$L_{simple}'$}{TEXT}}
\label{sec:lambda-exp}
\begin{figure}
  \centering
  \includegraphics[width=\textwidth]{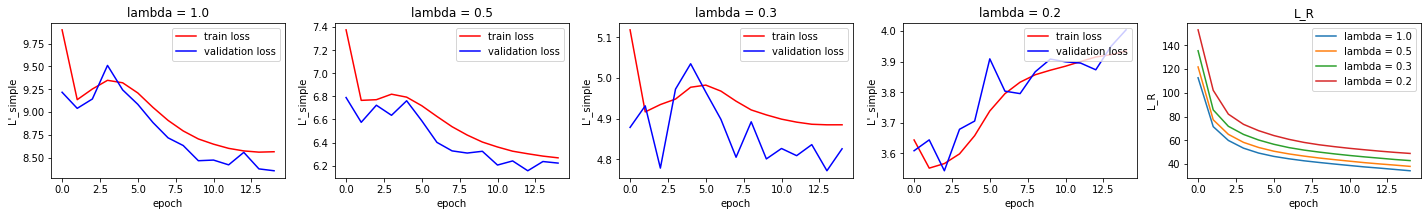}
  \caption{constant $L_R$ coefficient loss}
  \label{fig:const-lambda}
\end{figure}

\begin{figure}
  \centering
  \includegraphics[width=\textwidth]{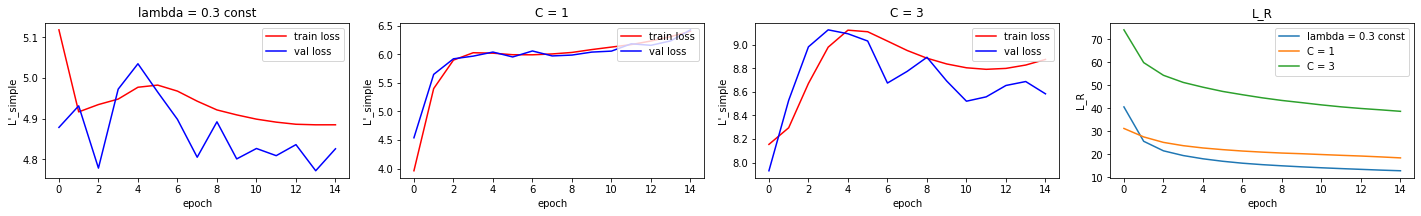}
  \caption{dynamic $L_R$ coefficient scheduling loss}
  \label{fig:dynamic-lambda}
\end{figure}

In this session, we explain our choice of rounding coefficient hyperparameter $\lambda$. We introduce the coefficient in the expectancy of balance the importance between $L_{simple}'$ and $L_R$ loss term in loss function. $\lambda \in \{1.0, 0.5, 0.3, 0.2\}$ are experimented on as shown in Figure 3. When $\lambda$ decreases, the model shows a decreasing trend in its $L_{simple}'$ loss (from 8.56 to 4.88) at the cost of increasing $L_R$ loss (from 34.4 to 48.9) as expected. In addition, setting $\lambda$ to lower than 0.3 shows an increase in $L_{simple}'$ loss, which means model is increasing the probability of outputting the ground truth sequence while not captioning the similarity between the predicted word and ground truth words. Though we find the final $L_{simple}'$ of $\lambda = 0.2$ (3.93) lower than that of $\lambda = 0.3$ (4.88), BLEU score of the model shows the opposite in Table 3. In our experiments on FLickr30+8k training, we compare the performance between $\lambda = 0.5$ and $\lambda = 0.3$ models when linear learning rate decay is enabled. The $\lambda = 0.5$ model early stops at around 8 epochs, and reports 0.2193 BLEU-4 score on validation set, which is worse than $\lambda = 0.3$ model's 0.2470 BLEU-4 score. 

Furthermore, we explore if a dynamic $\lambda$ scheduling method helps the model adjust $\lambda$ based on the ratio between $L_{simple}'$ and $L_R$. The dynamic scheduling method update $\lambda$ value after each gradient descend step as $\lambda = L_{simple}' / L_R * C$, where $C$ is a hyperparameter defining the relative weight kept between $L_{simple}'$ and $L_R$ throughout the training. To choose a reasonable $C$ value, we examine the $\lambda = 0.3$ experiment final epoch's $\frac{L_{simple}'}{L_R}$ value, which is around 2.80. Based on this observation, we take $C = 1$ and expect both $L_{simple}'$ and $L_R$ loss decrease at the same rate and reach $\lambda \approx 0.3$ in the final epoch. The same reasoning applies to the choice of $C = 3$, which aims at reaching $\lambda \approx 1.0$ in the final epoch. 

We plot the dynamic scheduled loss in Figure 4. The relative ratio between $L_{simple}'$ $L_R$ does not follow the dynamic ratio defined as expected. A possible explanation is that the loss is taken before gradient descent is performed, and evaluation of loss is based on the next batch sample, resulting in the plotted $\frac{L_{simple}'}{L_R}$ value not following the expected relative loss ratio. In both experiments, the model shows worse performance with dynamic $\lambda$ scheduling enabled compared with constant $\lambda$ scheduling. We conclude $\lambda = 0.3$ is a more suitable parameter for our CLIP-Diffusion-LM model.

\begin{minipage}[c]{0.49\textwidth}
    \centering
    \begin{tabular}{llll}
    \toprule
    $\lambda$ & learning rate & Flickr8k & Flickr30k \\
    \midrule
    1.0 & 5e-5 & 0.1599 & -\\
    0.5 & 5e-5 & 0.1614 & -\\
    0.5 & linear decay & - & 0.2193 \\
    0.3 & 5e-5 & 0.1549 & - \\
    0.3 & linear decay & - & 0.2470 \\
    0.2 & 5e-5 & 0.1550 & - \\
    \bottomrule
    \end{tabular}
    \captionsetup{type=table}\captionof{table}{BLEU-4 score of different constant $L_R$ coefficients}
    \label{tab:const-lambda}
\end{minipage}
\begin{minipage}[c]{0.5\textwidth}
    \centering
    \begin{tabular}{lllll}
    \toprule
    fusion method & Flickr8k & Flickr30+8k \\
    \midrule
    addition & 0.1033 & 0.1948 \\
    concatenation & 0.1549 & 0.2337 \\
    \bottomrule
    \end{tabular}
    \label{tab:fusion}
    \captionsetup{type=table}\captionof{table}{BLEU-4 score of different fusion methods}
\end{minipage}

\subsection{\texorpdfstring{$x_0$}{TEXT} prediction or \texorpdfstring{$x_{s_t-n}$}{TEXT} prediction}
\label{sec:x0-exp}
In Diffusion-LM proposed by Li et al. \cite{diffuselm}, model is trained to predict $x_0$. In this experiment, we examine the difference between $x_0$-prediction and $x_{s_t-n}$-prediction, where $x_{s_t-n}$-prediction trains the model to predict a constant $n = 100$ steps after the sampled latent embedding. In our experiments, the $x_0$-prediction method shows better BLEU-4 score (0.1549) and $L_{diffuseLM}$ loss (17.8) performance compared with $x_{s_{t-1}}$-prediction (0.1474 BLEU-4 score and 22.6 $L_{diffuseLM}$ loss). In addition, by following the $x_0$-prediction method, our model converges to a reasonable output in 5 diffusion stages. This inference step number is significantly less than that of an autoregressive encoder-decoder structure, which involves generation step number proportional to the output sequence length.

\subsection{fusion: concatenation or element-wise addition}
\label{sec:fusion-exp}
\begin{figure}
  \centering
  \includegraphics[width=\textwidth]{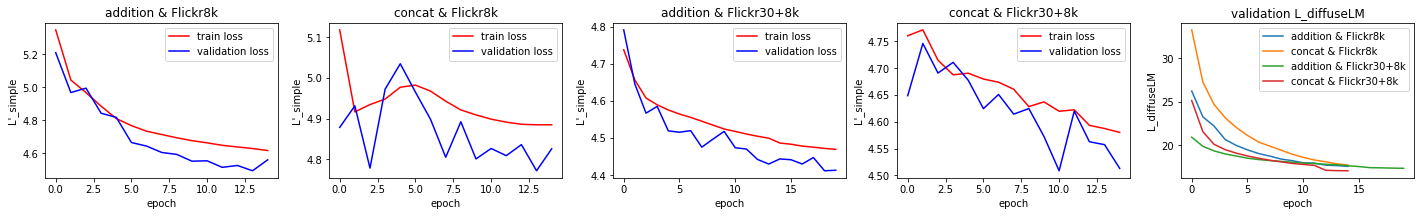}
  \caption{concatenation or addition fusion loss}
  \label{fig:fusion}
\end{figure}

We experiment with 2 fusion methods: concatenation or element-wise addition. Concatenation appends the CLIP features as additional tokens to the end of the caption embedding $x_t$. An additional segment embedding layer is used to distinguish CLIP feature tokens from caption embedding. In contrast, element-wise addition method use CLIP image feature as position embedding. CLIP image feature is element-wise added to caption word embedding in each timestep. In the case of classification-free training, the CLIP text feature is also element-wise added. In later explanation, element-wise addition is referred to as addition method for simplicity.

We conduct experiments by running both concatenation and addition fusion. Figure 5 and Table 4 show their performance on Flickr8k and Flickr 30+8k dataset. The addition method experience less violent fluctuation in $L_{simple}'$ loss, and converge to a lower loss before over-fitting. Extending the training dataset to use Flickr30+8k shows similar results. However, BLEU-4 scores of addition method (0.1033 on Flickr8k and 0.1948 on Flickr30+8k) are significantly lower than that of concatenation method (0.1549 on Flickr8k and 0.2337 on Flickr30+8k) in both datasets' training. As a result, concatenation method is chosen despite its higher converging loss and instability. 

\section{Conclusion}
We present the application of diffusion in the image caption task and prove its validity on Flickr8k and Flickr30k datasets. Particularly, we identify the importance of rounding term in loss function to help model converge, and introduce the adaptive ratio adjustment to balance its importance with other terms. There are various possible improvements to the model and training process:
\begin{itemize}
\item In various cases, the model fails to identify the correct object colour. For example, wrongly tag the object colour to be the background colour. We suggest 1) performing image data argumentation, which has been proven to improve the performance for VLP tasks  \cite{vilt}, and 2) investigating attention graph of the wrongly tagged object in image for possible causes.
\item The output text sequence from the model contains apparent grammar mistakes occasionally (e.g. missing subject and repeated words). Providing the model additional guidance on the output text grammar might help reduce grammar mistakes.
\item We limit the sequence generated by DistilBert to under 16 words due to the computation resource limit. As a result, the BLEU score suffers from the Brevity Penalty in BLEU metrics. Training model with relaxed length constrain might further improve the BLEU score.
\item In various cases models used in experiments with lower loss does not have a higher BLEU score, suggesting a better loss function is required to define model's performance.
\end{itemize}
We believe that analyzing and improving based on the above observations, using the diffusion language model as text generation step will achieve comparable or better performance than auto-regressive models in image captioning tasks.

\section*{Acknowledgments}
We thank Mu Li and Yi Zhu for sharing their insight in various models in vision and NLP field publicly online, Boyang Gu for providing advice in early stage of the research. The computation resource was supported by Imperial College London. 

\bibliographystyle{unsrt}  
\bibliography{references}

\end{document}